\documentclass[11pt,a4paper]{article}
\usepackage[hyperref]{emnlp2020}
\usepackage{times}
\usepackage{latexsym}

\usepackage{microtype}

\usepackage{natbib}
\usepackage[utf8]{inputenc} 
\usepackage[T1]{fontenc}    
\usepackage{hyperref}       
\usepackage{url}            
\usepackage{booktabs}       
\usepackage{amsfonts}       
\usepackage{nicefrac}       
\usepackage{microtype}      

\usepackage{graphicx}
\usepackage{caption}
\usepackage{subcaption}
\usepackage{wrapfig}

\usepackage{amsfonts}
\usepackage{amsmath}
\usepackage{amssymb}
\usepackage{amsthm}

\usepackage{algorithm}
\usepackage{algorithmic}

\usepackage{booktabs}
\usepackage{multirow}
\usepackage{tabularx}

\usepackage{xspace}
\usepackage{enumitem}

\usepackage{dblfloatfix}

\usepackage{balance}

\usepackage{color}

\aclfinalcopy 

\title{Summarizing Text on Any Aspects: \\ 
A Knowledge-Informed Weakly-Supervised Approach}

\author{
Bowen Tan$^1$,~~
Lianhui Qin$^2$,~~
Eric P. Xing$^{1,3}$,~~
Zhiting Hu$^{1,4}$\\
$^1$Carnegie Mellon University,~~ $^2$University of Washington,~~ $^3$Petuum Inc.,~~ $^4$UC San Diego\\
{\small 
{\tt \{btan2,epxing\}@andrew.cmu.edu, lianhuiq@cs.washington.edu, zhitinghu@gmail.com}
}
}

\date{}

\begin{document}
\maketitle
\begin{abstract}
Given a document and a target aspect (e.g., a topic of interest), aspect-based abstractive summarization attempts to generate a summary with respect to the aspect. Previous studies usually assume a small pre-defined set of aspects and fall short of summarizing on other diverse topics. In this work, we study summarizing on \emph{arbitrary} aspects relevant to the document, which significantly expands the application of the task in practice. Due to the lack of supervision data, we develop a new weak supervision construction method and an aspect modeling scheme, both of which integrate rich external knowledge sources such as ConceptNet and Wikipedia. Experiments show our approach achieves performance boosts on summarizing both real and synthetic documents given pre-defined or arbitrary aspects.\footnote{Code and data available at \url{https://github.com/tanyuqian/aspect-based-summarization}}
\end{abstract}





\section{Introduction}
Remarkable progresses have been made in generating \emph{generic} summaries of documents~\cite{nallapati2016abstractive,see2017get,narayan2018don}, partially due to the large amount of supervision data available. In practice, a document, such as a news article or a medical report, can span multiple topics or aspects. To meet more specific information need in applications such as personalized intelligent assistants, it is often useful to summarize a document with regard to a given aspect, i.e., \emph{aspect-based} summarization.

Recent research has explored the problem of aspect-based abstractive summarization~\cite{krishna2018generating,frermann2019inducing}. A key challenge of the task is the lack of direct supervision data containing documents paired with multiple aspect-based summaries. Previous studies have created synthetic data from generic news summarization corpora which have a small set of aspects (e.g., ``sports'', ``health'' and other 4 aspects in ~\cite{frermann2019inducing}). As a result, models trained on these data tend to be restricted to the pre-defined set and fall short of summarizing on other diverse aspects.

This paper aims to go beyond pre-defined aspects and enable summarization on arbitrary aspects relevant to the document.
The arbitrary aspect may not be explicitly mentioned but only implicitly related to portions of the document, and it can be a new aspect not seen during training.
To this end, we develop a new approach that integrates rich external knowledge in both aspect modeling and weak supervision construction. Specifically, we derive weak supervisions from a generic summarization corpus, where the ConceptNet knowledge graph~\cite{speer2017conceptnet} is used to substantially expand the aspect scope and enrich the supervisions. 
To assist summarization model to better understand an aspect, especially a previously unseen one, we augment the model inputs with rich aspect-related information extracted from Wikipedia.

Our approach is compatible with any neural encoder-decoder architectures. In this work, we use the large pre-trained BART model~\cite{lewis2019bart} and fine-tune with the proposed method. 
Experiments on real news articles show our approach achieves performance boosts over existing methods. 
When adapting to the previous synthetic domain, the BART model after fine-tuning with our weak supervisions becomes substantially more data efficient, and outperforms previous best-performing systems greatly using only 0.4\% training examples.






\begin{figure*}[t]
    \centering
    \includegraphics[width=1.0\textwidth]{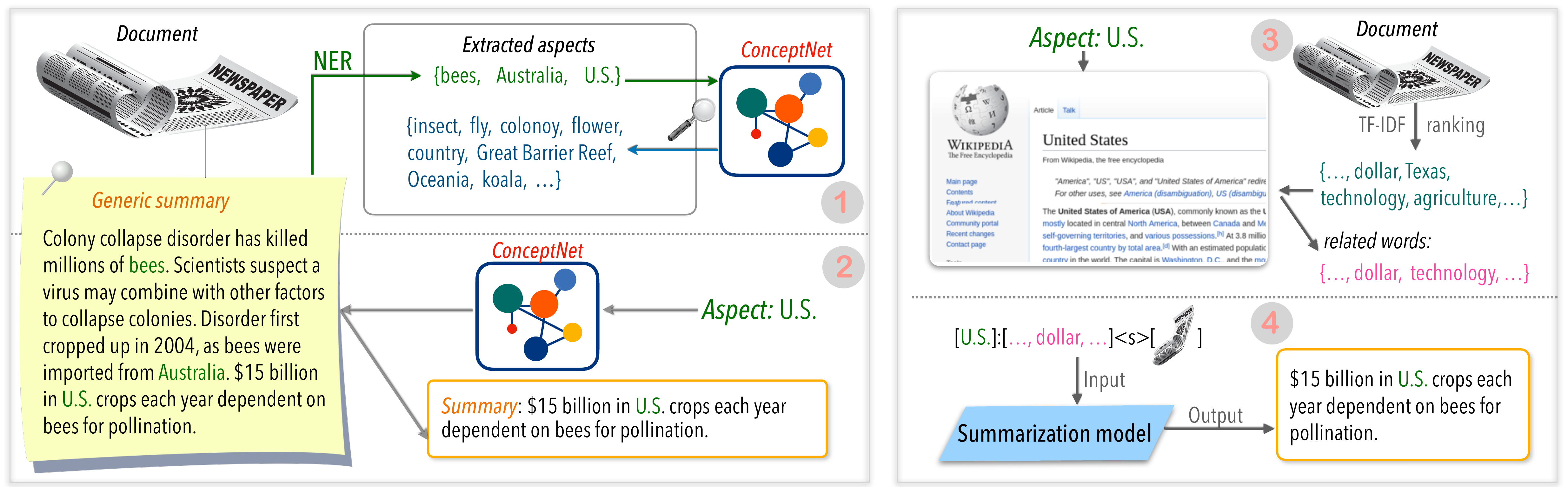}
    \vspace{-16pt}
    \caption{
    Illustration of our approach. {\bf Left:} Constructing weak supervisions using ConceptNet, including (1) extracting aspects and (2) synthesizing aspect-based summaries. {\bf Right:} Augmenting aspect information, including (3) identifying aspect related words in the document using Wikipedia and (4) feeding both aspect and related words into summarization model.
    }
    \label{fig:model}
    \vspace{-15pt}
\end{figure*}

\vspace{-5pt}
\section{Related Work}\label{sec:related}
\vspace{-5pt}

Aspect-based summarization as an instance of controllable text generation~\citep{hu2017controllable,ficler2017controlling} offers extra controllability compared to generic summarization to ensure concise summaries of interest.
Early work has studied topic-aware summarization in the multi-document setting, with (typically small) datasets containing multiple documents tagged with a relevant topic~\citep{dang2005overview,conroy2006topic}. For single-document aspect-based summarization, \emph{extractive} methods were used to extract related key sentences/words from the document~\cite{lin2000automated}. 
Our work studies \emph{abstractive} aspect-based summarization that generates summaries. \citet{deutsch2019summary} studied a sub-task of learning to select information in documents that should be included in the summary. Recent work~\cite{frermann2019inducing,krishna2018generating} on the problem synthesized training data that use news categories as the aspects and thus have a small pre-defined set of aspects available. We aim to enable summarization on any aspects, and develop new weak supervisions by integrating rich external knowledge.  

Aspect-based summarization has also been explored in the customer reviews domain~\cite{hu2004mining}, where product aspects, customer sentiment, and sometimes textual summaries are extracted~\cite{popescu2007extracting,wang2016neural,angelidis2018summarizing}. Query-based summarization produces a summary in response to a natural language query/question~\cite{daume2006bayesian,liu2012query,xie2020conditional} which differs from abstract aspects.

Incorporating knowledge through weak supervision has primarily been studied in classification or extraction problems~\citep{hu2016harnessing,peng2016event,ratner2017snorkel}. For example, \citep{hu2016harnessing} creates soft labels from a logical-rule enhanced teacher model to train neural classifiers. This work explores weak supervisions in the generation setting. Automatic creation of data supervisions also links our work to text data augmentation in either heuristic-based~\citep{wei2019eda} or automated manner~\citep{sennrich2016improving,hu2019learning}. This work embeds rich structured knowledge in the data synthesis process.

\vspace{-5pt}
\section{Approach}\label{sec:method}
\vspace{-5pt}

Given a document and an aspect which can be a word or a phrase, the task aims to generate a summary that concisely describes information in the document that is relevant to the aspect.
%
We present our approach that enables a neural summarization model to summarize on any aspects. The aspect can be any words relevant to (but not necessarily occurring in) the document. Our approach incorporates rich external knowledge sources, including ConceptNet for enriching weak supervisions in training (sec~\ref{sec:method:supervision}) and Wikipedia for advising the document-aspect relation to improve comprehension (sec~\ref{sec:method:aspects}).
Figure~\ref{fig:model} shows an overview of our approach.

An advantage of our approach is that it is compatible with any neural summarization architectures, such as the popular encoder-decoders. This enables us to make use of the large pre-trained network BART~\cite{lewis2019bart}, on which we apply our approach for fine-tuning and improved inference.

\vspace{-5pt}
\subsection{Knowledge-enriched Weak Supervisions}
\label{sec:method:supervision}

Usually no direct supervision data is available. We start with a generic summarization corpus. Specifically, in this work we use the CNN/DailyMail \cite{hermann2015teaching} which consists of a set of \textit{(document, summary)} pairs. Our approach constructs weakly supervised examples by automatically extracting potential aspects and synthesizing aspect-based summaries from the generic summary. Each resulting aspect and its aspect-based summary are then paired with the document for training.

\vspace{-5pt}
\paragraph{Extracting Aspects}
Given a generic summary, we want to extract as many aspects as possible so that the summarization model can see sufficient examples during training. On the other hand, the aspects must be relevant to the generic summary to facilitate synthesizing appropriate summary in the next step. To this end, we first apply a named entity recognition (NER) model\footnote{https://spacy.io/models/xx} to extract a set of entities mentioned in the generic summary. These entities serve as a seed set of aspects. We then augment the seed set by collecting each entity's neighbor concepts on the ConceptNet knowledge graph, as these concepts are semantically closely related to the entity (and thus the generic summary). For example, in Figure~\ref{fig:model}(1), ``insect'' is a new aspect from ConceptNet given the seed entity ``bees''.

\vspace{-5pt}
\paragraph{Synthesizing Aspect-based Summaries}
For each aspect, we synthesize a specific summary by extracting and concatenating all relevant sentences from the generic summary. We make use of ConceptNet in a similar way as above. Specifically, a sentence is considered relevant if it mentions the aspect or any of its neighbors on ConceptNet.

The use of ConceptNet greatly augments the supervisions in terms of both the richness of aspects and the informativeness of respective summaries.

\vspace{-5pt}
\subsection{Knowledge-aided Aspect Comprehension}\label{sec:method:aspects}

The summarization model is required to precisely locate information in the document that matches the desired aspect. Such comprehension and matching can be challenging, especially with only noisy weak supervisions during training. Our approach facilitates the inference by informing the model with pre-computed document-aspect relations. 

Concretely, we extract words from the document which are most \emph{related} to the aspect (more details below), and feed those words into the model together with the aspect and document. In this way, the model is advised which parts of the document are likely to be aspect-related.
For the BART architecture, we use an input format as:
\begin{center}
\texttt{[aspect]:[related words]<s>[doc]}
\end{center}
where \texttt{<s>} is a special token for separation.

To determine the related words, the intuition is that the words should be describing or be associated with the aspect. We use the Wikipedia page of the aspect for filtering the words. Besides, we want to select only salient words in the document for a concise summary. Thus, we first rank all words in the document by TF-IDF scores, and select top words that occur in the aspect's Wikipedia page\footnote{We select $\leq 10$ words. If the Wikipedia API does not find any page of the aspect, the related word is set to empty.}. 

\vspace{-5pt}
\section{Experiments}\label{sec:exp}
\vspace{-5pt}




\paragraph{Setup}

We construct weak supervisions from 100K out of 280K (doc, summary) pairs in the training set of the CNN/DailyMail dataset~\cite{hermann2015teaching}.
%
%
We use the CNN/DailyMail-pretrained BART \cite{lewis2019bart} provided by Fairseq~\cite{ott2019fairseq} as our base summarization model, and fine-tune with our approach implemented using Texar~\citep{hu2019texar}. 
We use Adam optimizer with
an initial learning rate of 3e-5,
and beam search decoding with a width of 4.

\vspace{-5pt}
\subsection{Studies on Synthetic Domain}
We first study on the synthetic data, {\bf MA-News}, introduced in~\cite{frermann2019inducing}. 
Although its aspects are restricted to only 6 coarse-grained topics, the synthetic domain facilitates automatic evaluation, providing a testbed for (1) comparison with the previous models and (2) studying the generalization ability of our weak-supervision approach when adapting to the new domain.

Specifically, MA-News is synthesized from CNN/DailyMail by interleaving paragraphs of original documents belonging to different aspects. The assembled document is paired with each component's aspect and generic summary to form an aspect-based summary instance. The dataset has 280K/10K/10K examples in 
train/dev/test sets, respectively, and contains 6 pre-defined aspects including \{``sport", ``health", ``travel", ``news", ``science technology", ``tv showbiz"\}. 


\begin{table}[t]
\small
    \centering
    \begin{tabular}{@{}rlll@{}}
\toprule
{\bf Models}          & {\bf R-1} & {\bf R-2} & {\bf R-L} \\ \cmidrule{1-4}
Lead-3~(\citeyear{frermann2019inducing}) & 21.50   & 6.90    & 14.10   \\
PG-Net~(\citeyear{see2017get})                 & 17.57   & 4.72    & 15.94   \\
SF~(\citeyear{frermann2019inducing}) & 28.02   & 10.46   & 25.36   \\ \cmidrule{1-4}
BART MA-News-Sup 280K & {41.90}   & {20.46}   & {39.06}   \\
BART MA-News-Sup 1K  & 24.58          & 8.82           & 22.74          \\
\cmidrule{1-4}
BART Weak-Sup (Ours) & 28.56   & 10.53    & 25.93 \\
+ MA-News-Sup 1K (Ours) & {35.62}  & {15.80} & { 33.01} \\
\bottomrule
\end{tabular}
\vspace{-2pt}
\caption{
Results (ROUGE) on the MA-News test set. 
The results of \texttt{Lead-3}, \texttt{PG-Net} and \texttt{SF} are from \cite{frermann2019inducing}, where \texttt{SF} is the previous best model. 
Our approach trains with only weak supervisions (sec~\ref{sec:method:supervision}) or with additional 1K MA-News supervised training data.
}
\label{tab:syn-performance}
\end{table}

\begin{table}[t]
\small
    \centering
    \begin{tabular}{@{}rlll@{}}
\toprule
       {\bf Models}        &  {\bf R-1}        &  {\bf R-2}        & {\bf R-L}        \\ \midrule
Weak-Sup only     & 28.56          & 10.53          & 25.93          \\ \cmidrule{1-4}
MA-News-Sup 1K  & 24.58          & 8.82           & 22.74          \\
+ Weak-Sup & \textbf{35.62} & \textbf{15.80} & \textbf{33.01} \\ \cmidrule{1-4}
MA-News-Sup 3K  & 29.13          & 11.89          & 27.02          \\
+ Weak-Sup & \textbf{37.17} & \textbf{16.84} & \textbf{34.40} \\ \cmidrule{1-4}
MA-News-Sup 10K & 39.49          & 18.71          & 36.67          \\
+ Weak-Sup & \textbf{39.82} & \textbf{18.81} & \textbf{36.92} \\ \bottomrule
\end{tabular}
\vspace{-2pt}
\caption{
Fine-tuning BART on the synthetic domain, evaluated on MA-News test set. 
\texttt{Weak-Sup~only} trains BART only with our weak supervisions. \texttt{MA-News-Sup~1K} trains with 1K MA-News supervised examples. \texttt{+Weak-Sup} trains first with weak supervisions and then supervisedly on MA-News.
}
\label{tab:warm-start}
\vspace{-15pt}
\end{table}

\vspace{-5pt}
\paragraph{Comparisons with previous methods}
We first compare our approach with the previous summarization models, as shown in Table~\ref{tab:syn-performance}. 
{\bf (1)} 
In the first block,
\texttt{SF} is the best model in~\cite{frermann2019inducing} with a customized neural architecture and is trained with the full MA-News training set.  
{\bf (2)} In the second block, we also train the large BART model with the MA-News training set, either using the full 280K instances or only 1K instances. BART trained with the full set unsurprisingly shows much better results than \texttt{SF}, yet the one with the 1K subset falls behind \texttt{SF}. 
{\bf (3)} The third block evaluates our method. \texttt{BART Weak-Sup} is fine-tuned only with our weak supervisions (sec~\ref{sec:method}). Even without using any direct supervision examples in MA-News, the model performs slightly better than \texttt{SF}. More interestingly, by further using only 1K MA-News instances to continue fine-tuning the model, we achieve performance boosts compared to both \texttt{SF} and \texttt{BART~MA-News-Sup~1K}. This shows our proposed knowledge-informed method provides rich information that helps with the task.
\begin{figure}[t]
    \centering
    \vspace{-15pt}
    \includegraphics[width=0.45\textwidth]{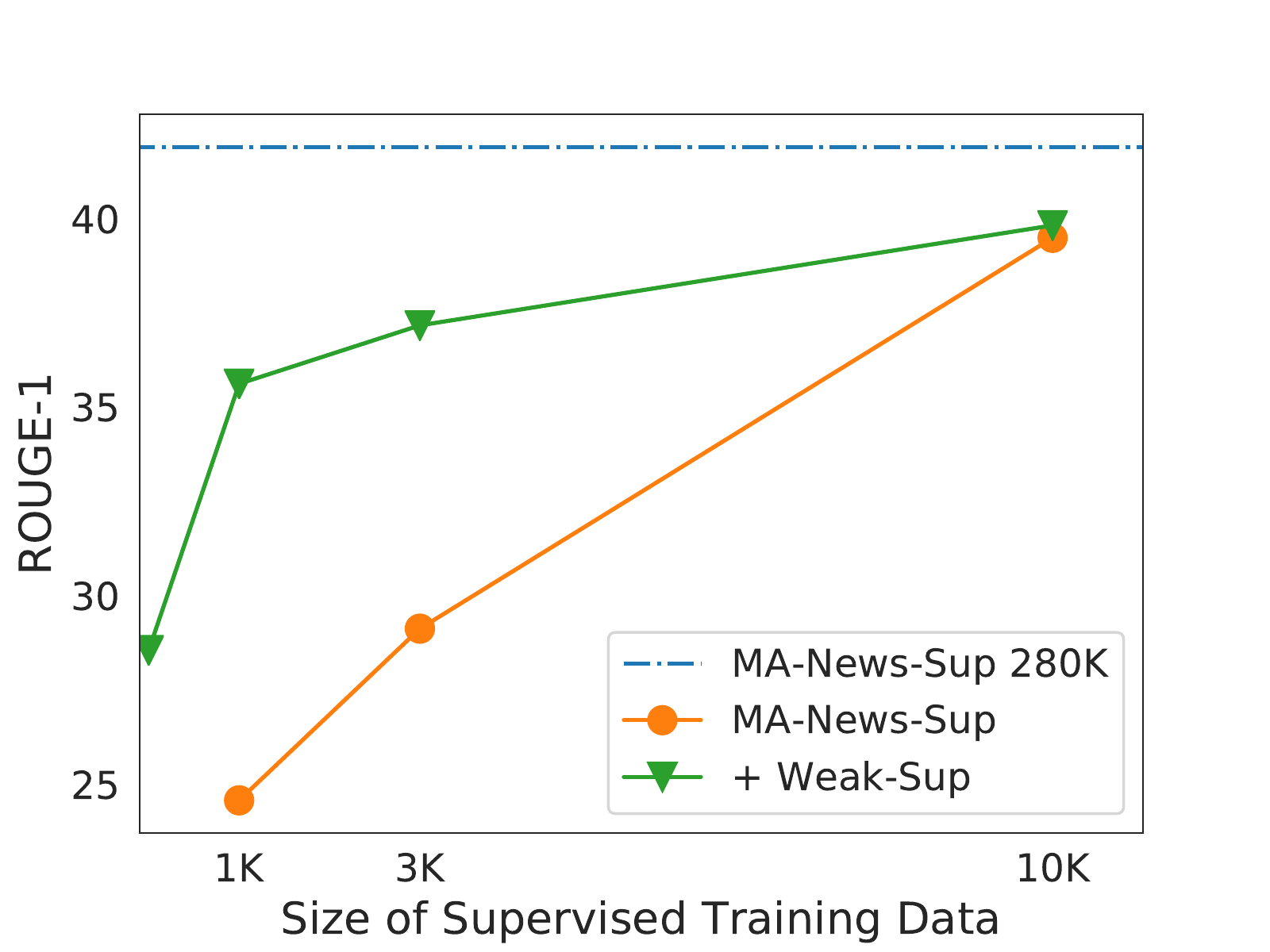}
    \vspace{-6pt}
    \caption{
    Visualizing the ROUGE-1 results in Table~\ref{tab:warm-start}. The green dashed line marks the performance of BART fine-tuned on the whole MA-News training set. 
    }
    \label{fig:sample_efficiency}
\end{figure}

\vspace{-5pt}
\paragraph{Efficiency of adapting to the domain}
We continue to study how our weakly supervised method can help with efficient adaptation of BART to the synthetic domain. As shown in Table~\ref{tab:warm-start}, by fine-tuning BART using more MA-News training data (i.e., \texttt{MA-News-Sup 1K}, \texttt{3K}, and \texttt{10K}), the test performance improves reasonably, as is also shown by the blue curve in Figure~\ref{fig:sample_efficiency}. 
However, if we add our proposed weak supervisions 
(i.e., \texttt{+Weak-Sup}), 
the performance improves much faster, as is also shown by the orange curve in the figure. The enhanced data efficiency validates the effectiveness of the weakly supervised method.

\begin{table}[t]
\small
    \centering
    \begin{tabular}{@{}rlll@{}}
\toprule
        {\bf Models}     &  {\bf Accu.} &  {\bf Info.} & {\bf Fluency} \\ \midrule
MA-News-Sup 280K   & 2.19     & 3.44  & 4.44    \\ [2pt]
Weak-Sup (Ours)    &  {\bf 4.59}     &  {\bf 4.36}           & {\bf 4.87}    \\ [2pt]
+ MA-News 3K (Ours)& 4.14     & 4.07            & 4.80    \\ \bottomrule
\end{tabular}
\vspace{-6pt}
    \caption{Human evaluation using 5-point Likert scale. \texttt{MA-News 280K} trains BART with the whole MA-News set. \texttt{Weak-Sup} trains with our weak supervisions. \texttt{+MA-News 3K} further fine-tunes with 3K MA-News instances.}
    \label{tab:human-eval}
    \vspace{-20pt}
\end{table}

\begin{figure}
    \vspace{-15pt}
    \centering
    \includegraphics[width=0.48\textwidth]{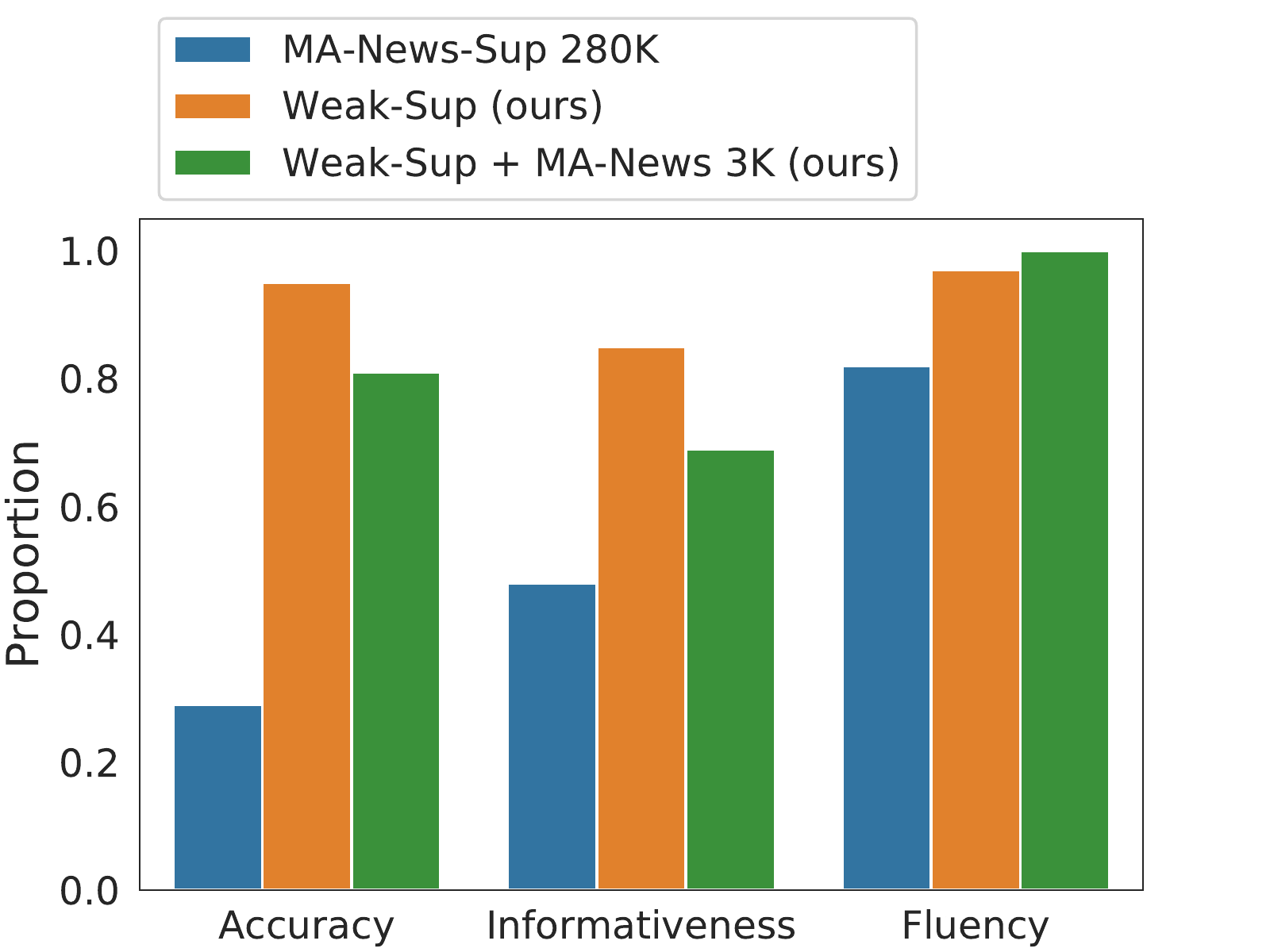}
    \vspace{-20pt}
    \caption{
    Proportions of model outputs that get a human score $\geq$ 4. For example, around 95\% of summaries by \texttt{Weak-Sup (ours)} are scored 4 or 5 in terms of accuracy. 
    }
    \label{fig:human_high_scores}
\end{figure}

\begin{table*}[h]
    \centering
    \small
    \begin{tabular}{p{2\columnwidth}@{}l@{}}
\toprule
\textbf{Document} 
In an exclusive interview with Breitbart News, {\color{cyan} Republican presidential nominee Donald Trump blasted Bill \textbf{Clinton}’s suggestion that the United States use \textbf{Syrian refugees} to rebuild Detroit. The populist \textbf{billionaire} denounced \textbf{Clinton}’s suggested proposal as “\textbf{crazy}” and “unfair” to \textbf{American} workers who are already living there and are in need of \textbf{jobs}.  “It’s very unfair to the \textbf{people} that are living there.} I think it’s crazy,” Trump told Breitbart on Thursday. “I mean, these people 
... ...
“There are plenty of people in Detroit who you could almost look at as refugees,” Carson said. “I mean, we need to take care of our own people. We need to create jobs for them. ” { \color{magenta} \textbf{Clinton}’s suggestion that the U. S. ought to give Detroit jobs to foreign \textbf{refugees} came during a February discussion at the Clinton Global Initiative with  Chobani \textbf{billionaire} and mass migration enthusiast, Hamdi Ulukaya.} “The truth is that the big loser in this over the long run is
... ...
a pretty good deal. ” { \color{magenta} During the discussion, \textbf{Clinton} praised Ulukaya for his efforts to fill his \textbf{yogurt} plants with \textbf{imported} foreign \textbf{refugees}. Ulukaya suggested that the U. S. ought to be taking in more \textbf{refugees} and said that he was “proud” of Turkey’s decision to accept 2 million \textbf{Syrian refugees}. Ulukaya told \textbf{Clinton} that \textbf{Syrian refugees} “bring \textbf{flavors} to the community just like in  …     Twin Falls, [Idaho]” where Ulukaya’s \textbf{yogurt} factory is based. } Clinton’s controversial suggestion that 
... ...
millions of more illegal immigrants, thousands of more violent crimes, and total chaos and lawlessness. { \color{orange} \textbf{According} to Pew \textbf{polling} data, Hillary Clinton’s plan to expand immigration is opposed by at least 83 \textbf{percent} of the \textbf{American} \textbf{electorate}  —   \textbf{voters} whom Clinton has \textbf{suggested} are racist for opposing   immigration. \textbf{According} to a September 2015 Rasmussen survey, 85 \textbf{percent} \textbf{black} \textbf{voters} oppose Clinton’s refugee agenda to admit more than 100, 000 Middle Eastern refugees —  with less than one \textbf{percent} of \textbf{black voters} (. 56 \textbf{percent}) in \textbf{favor} of her refugee plan.}
 \\ \midrule
 
\textbf{Aspect:} { \color{cyan} Donald Trump} \\ [2pt]
\textbf{Summary:} Presidential nominee Donald Trump calls suggestion that the u.s. use Syrian refugees to rebuild Detroit "crazy" and "unfair" to American workers who are already living there and in need of jobs. \\ \midrule

\textbf{Aspect:} {\color{magenta} Hamdi Ulukaya} \\ [2pt]
\textbf{Summary:} Chobani billionaire and mass migration enthusiast, Hamdi Ulukaya, suggested that the U.S. should take in more refugees to fill jobs like in his yogurt plant in Twin Falls, Idaho, where his factory is based.  \\ \midrule

\textbf{Aspect:} {\color{orange} vote} \\ [2pt]
\textbf{Summary:} Polls show that at least 83 percent of the U.S. electorate is opposed to expanding immigration and that 85 percent of black voters oppose the plan to admit more than 100,000 middle eastern refugees to the country.  \\

\bottomrule
\end{tabular}
\vspace{-6pt}
\caption{Generated summaries of a document on different aspects. Document content relevant to specific aspects is highlighted in respective colors. ``Related words'' identified through Wikipedia (sec~\ref{sec:method:aspects}) are highlighted in {\bf bold}.
}
\vspace{-6pt}
\label{tab:full_example}
\end{table*}

\subsection{Summarizing Real News on Any Aspects}

We next study summarization of a document on arbitrary aspects.
To evaluate the generalization of the methods, we test on real news articles from the \emph{All The News} corpus~\cite{allthenews} where we randomly extract 50 articles from different publications other than CNN (so that no articles are included in the weak supervision). We ask human annotators to label an arbitrary relevant aspect for each article. We then collect aspect-based summaries by the models, and present each to 3 annotators to rate. 

As in previous work~\citep{kryscinski2019neural,frermann2019inducing}, the criteria include \emph{accuracy} (coherence between the aspect and the summary
), \emph{informativeness} (factual correctness and relevance of the summary with regard to the document), and \emph{fluency} (language quality of individual sentences and the whole summary). The Pearson correlation coefficient of human scores is 0.51, showing moderate inter-rater agreement. 
Table~\ref{tab:human-eval} shows the averaged scores, and Figure~\ref{fig:human_high_scores} shows the proportions of model outputs receiving high scores in terms of the three criteria. We can see our weakly supervised method performs best. The model trained on the 280K MA-News examples, though performs well on the MA-News test set (Table~\ref{tab:syn-performance}), fails to generalize to the broader set of diverse aspects, showing the importance of introducing rich knowledge in supervisions and inference process for generalization.
Interestingly, fine-tuning our model with 3K MA-News instances results in inferior performance, showing the previous synthetic data with limited aspects could restrict generalization to other aspects.


Table~\ref{tab:full_example} shows example summaries by our \texttt{Weak-Sup} model. Given an arbitrary aspect (e.g., an entity or a word), the model correctly identifies the related portions in the document and generates a relevant short summary. It is also noticeable that our approach identifies meaningful ``related words'' using Wikipedia as described in sec~\ref{sec:method:aspects}, which help with precise summarization.

\vspace{-5pt}
\section{Conclusions}
\vspace{-5pt}

This paper studies the new problem of summarizing a document on arbitrary relevant aspects. To tackle the challenge of lacking supervised data, we have developed a new knowledge-informed weakly supervised method that leverages external knowledge bases.
The promising empirical results motivate us to explore further the integration of more external knowledge and other rich forms of supervisions (e.g., constraints, interactions, auxiliary models, adversaries)~\citep{hu2020learning,ziegler2019fine} in learning. We are also interested in extending the aspect-based summarization in more application scenarios (e.g., summarizing a document corpus).

\vspace{-20pt}

\balance
\bibliographystyle{acl_natbib}
\bibliography{anthology,emnlp2020}


\appendix


\onecolumn

\section{More Experimental Details}

We use Adam optimizer with
$\beta = (0.9, 0.999), \epsilon = 10^{-8}$, 
a weight decay of 0.01, and 
an initial learning rate of 3e-5.
For generation, we use beam search decoding with a width of 4 and a length penalty of 2.
All experiments are conducted on 4 GTX 1080Ti GPUs.

\section{More Generation Examples}
We provide more generated summaries from our weakly supervised model.

\begin{table*}[h]
    \centering
    \small
    \begin{tabular}{p{\columnwidth}@{}l@{}}
\toprule
\textbf{Document} 
In an exclusive interview with Breitbart News, { \color{cyan} Republican presidential nominee Donald Trump blasted Bill Clinton’s suggestion that the United States use Syrian refugees to rebuild Detroit. The populist billionaire denounced Clinton’s suggested proposal as “crazy” and “unfair” to American workers who are already living there and are in need of   jobs.  “It’s very unfair to the people that are living there. I think it’s crazy,” Trump told Breitbart on Thursday.} “I mean, these people are getting started —  I think it’s a very, very hard place to get your start. ” “We shouldn’t have them [i. e. Syrian refugees] in the country,” Trump added. “We don’t know who these people are. We have no idea. This could be the all time great Trojan horse. We have no idea who they are. The whole thing is ridiculous. Number one: we should build safe zones over in Syria, that’s what we should have, and we should have the Gulf states fund them. It’s just crazy. We ought to be building safe zones in Syria and not taking these people in —   whether it’s Detroit or anywhere else. ”   and former GOP presidential contender Ben Carson echoed Trump’s sentiment in a Friday interview on Breitbart News Daily on SiriusXM Patriot Channel 125. Carson explained that “we need to take care of our own people” and noted that the policies of Democrat politicians have turned many Americans living in Detroit into refugees in their own country. “There are plenty of people in Detroit who you could almost look at as refugees,” Carson said. “I mean, we need to take care of our own people. We need to create jobs for them. ” { \color{magenta} Clinton’s suggestion that the U. S. ought to give Detroit jobs to foreign refugees came during a February discussion at the Clinton Global Initiative with Chobani billionaire and mass migration enthusiast, Hamdi Ulukaya.} “The truth is that the big loser in this over the long run is going to be Syria. This [i. e. the Syrian migrant crisis] is an enormous opportunity for Americans,” Clinton said in February.  “Detroit has 10, 000 empty, structurally sound houses  —  10, 000. And lot of jobs are to be had repairing those houses. Detroit just came out of bankruptcy and the mayor’s trying to do an innovative sort of urban homesteading program there. But it just gives you an example of what could be done. And I think any of us who have ever had any personal experience with either Syrian Americans or Syrian refugees think it’s a pretty good deal. ” During the discussion, { \color{magenta} Clinton praised Ulukaya for his efforts to fill his yogurt plants with imported foreign refugees. Ulukaya suggested that the U. S. ought to be taking in more refugees and said that he was “proud” of Turkey’s decision to accept 2 million Syrian refugees. Ulukaya told Clinton that Syrian refugees “bring flavors to the community just like in  …     Twin Falls, [Idaho]” where Ulukaya’s yogurt factory is based.} Clinton’s controversial suggestion that U. S. give Detroit jobs to imported foreign refugees has received virtually no media coverage outside of Breitbart News. During his Wednesday immigration policy address, Trump challenged the media to begin asking Hillary Clinton to explain her “radical” immigration platform. The media needs to begin demanding to hear Hillary Clinton’s answer on how her policies will affect Americans and their security  …     These are matters of    for our country and its people, and we deserve answers from Hillary Clinton  …     What we do know, despite the total lack of media curiosity, is that Hillary Clinton promises a radical amnesty combined with a radical reduction in immigration enforcement. The result will be millions of more illegal immigrants, thousands of more violent crimes, and total chaos and lawlessness. { \color{orange} According to Pew polling data, Hillary Clinton’s plan to expand immigration is opposed by at least 83 percent of the American electorate  —   voters whom Clinton has suggested are racist for opposing   immigration. According to a September 2015 Rasmussen survey, 85 percent black voters oppose Clinton’s refugee agenda to admit more than 100, 000 Middle Eastern refugees —  with less than one percent of black voters (. 56 percent) in favor of her refugee plan.}
 \\ \midrule
 
\textbf{Aspect:} { \color{cyan} Donald Trump} \\ [2pt]
\textbf{Summary:} Presidential nominee Donald Trump calls suggestion that the u.s. use Syrian refugees to rebuild Detroit "crazy" and "unfair" to American workers who are already living there and in need of jobs. \\ \midrule

\textbf{Aspect:} {\color{magenta} Hamdi Ulukaya} \\ [2pt]
\textbf{Summary:} Chobani billionaire and mass migration enthusiast, Hamdi Ulukaya, suggested that the U.S. should take in more refugees to fill jobs like in his yogurt plant in Twin Falls, Idaho, where his factory is based.  \\ \midrule

\textbf{Aspect:} {\color{orange} vote} \\ [2pt]
\textbf{Summary:} Polls show that at least 83 percent of the U.S. electorate is opposed to expanding immigration and that 85 percent of black voters oppose the plan to admit more than 100,000 middle eastern refugees to the country.  \\

\bottomrule
\end{tabular}
\end{table*}

\begin{table*}[h]
    \centering
    \small
    \begin{tabular}{p{\columnwidth}@{}l@{}}
\toprule
\textbf{Document} 
{ \color{cyan} the palestinian authority officially became the 123rd member of the international criminal court on wednesday, a step that gives the court jurisdiction over alleged crimes in palestinian territories. the formal accession was marked with a ceremony at the hague, in the netherlands, where the court is based. the palestinians signed the icc's founding rome statute in january, when they also accepted its jurisdiction over alleged crimes committed "in the occupied palestinian territory, including east jerusalem, since june 13, 2014."} later that month, the icc opened a preliminary examination into the situation in palestinian territories, paving the way for possible war crimes investigations against israelis. as members of the court, palestinians may be subject to counter-charges as well. {\color{magenta} israel and the united states, neither of which is an icc member, opposed the palestinians' efforts to join the body. but palestinian foreign minister riad al-malki, speaking at wednesday's ceremony, said it was a move toward greater justice.} {\color{magenta} "as palestine formally becomes a state party to the rome statute today, the world is also a step closer to ending a long era of impunity and injustice," he said, according to an icc news release. "} indeed, today brings us closer to our shared goals of justice and peace." judge kuniko ozaki, a vice president of the icc, said acceding to the treaty was just the first step for the palestinians. "as the rome statute today enters into force for the state of palestine, palestine acquires all the rights as well as responsibilities that come with being a state party to the statute. these are substantive commitments, which cannot be taken lightly," she said. rights group human rights watch welcomed the development. "governments seeking to penalize palestine for joining the icc should immediately end their pressure, and countries that support universal acceptance of the court's treaty should speak out to welcome its membership," said balkees jarrah, international justice counsel for the group. "what's objectionable is the attempts to undermine international justice, not palestine's decision to join a treaty to which over 100 countries around the world are members." in january, when the preliminary icc examination was opened, israeli prime minister benjamin netanyahu described it as an outrage, saying the court was overstepping its boundaries. the united states also said it "strongly" disagreed with the court's decision. "as we have said repeatedly, we do not believe that palestine is a state and therefore we do not believe that it is eligible to join the icc," the state department said in a statement. it urged the warring sides to resolve their differences through direct negotiations. "we will continue to oppose actions against israel at the icc as counterproductive to the cause of peace," it said. but the icc begs to differ with the definition of a state for its purposes and refers to the territories as "palestine." while a preliminary examination is not a formal investigation, it allows the court to review evidence and determine whether to investigate suspects on both sides. prosecutor fatou bensouda said her office would "conduct its analysis in full independence and impartiality." the war between israel and hamas militants in gaza last summer left more than 2,000 people dead. the inquiry will include alleged war crimes committed since june. the international criminal court was set up in 2002 to prosecute genocide, crimes against humanity and war crimes. cnn's vasco cotovio, kareem khadder and faith karimi contributed to this report. \\ \midrule
\textbf{Aspect:} {\color{cyan} palestinian} \\
\textbf{Summary:} palestinian foreign minister: "the world is also a step closer to ending a long era of impunity and injustice". the palestinian authority becomes the 123rd member of the international criminal court. the move gives the court jurisdiction over alleged crimes committed in palestine since june.  \\
 \midrule

\textbf{Aspect:} {\color{magenta} israel} \\
\textbf{Summary:} israel and the u.s. opposed the palestinians' efforts to join the court, which is based in the netherlands. palestinian foreign minister: "today brings us closer to our shared goals of justice and peace" the court has jurisdiction over alleged crimes committed in the occupied territories since june.  \\
\bottomrule
\end{tabular}
\vspace{-6pt}
\end{table*}

\begin{table*}[h]
    \centering
    \small
    \begin{tabular}{p{\columnwidth}@{}l@{}}
\toprule
\textbf{Document} 
 Around 1980, Bill Gates gave Microsoft, the company he founded, a : ”A computer on every desk and in every home. ”’ ”But Microsoft CEO Satya Nadella, who  in 1992 and rose to the top job in 2014, thinks Gates’ famous mission had a big flaw.” ’”When I joined the company in 1992, we used to talk about our mission as putting a PC in every home, and by the end of the decade we have done that, at least in the developed world,” Nadella told  published on Monday. ”It always bothered me that we confused an enduring mission with a temporal goal. ”’ ”In other words, Nadella is saying that Gates’ vision for the future of Microsoft had a logical stopping point and did not consider what the company’s direction would be once the goal was achieved. {\color{cyan} In the 2000s, under former CEO Steve Ballmer, Microsoft became better known for its efforts to  than for innovating.” ’Nadella believes in making Microsoft more driven by a sense of purpose  —   in 2015, he said  was ”to empower every person and every organization on the planet to achieve more.” And he has encouraged the company .’ ”Under Nadella, Microsoft has taken the focus off Windows and the PC and pinned its hopes to the rise of its Azure and Office 365   products,  as the company’s older businesses stagnate.} While this approach hasn’t translated to huge revenue growth, it has revitalized the company’s image.” {\color{magenta} ”Gates told USA Today that he enjoys working with Nadella, serving his   successor as a special adviser and helping guide the company’s investments in   technology as it competes with Apple, Google, and Amazon.”}  \\ \midrule


\textbf{Aspect:} { \color{cyan} innovation} \\
\textbf{Summary:} In the 2000s, under former CEO Steve Ballmer, Microsoft became better known for its efforts to  than for innovating. under Nadella, Microsoft has taken the focus off Windows and the PC and pinned its hopes to the rise of its Azure and Office 365 products.   \\ \midrule

\textbf{Aspect:} { \color{magenta} apple} \\
\textbf{Summary:} Gates says he enjoys working with Nadella, serving as a special adviser and helping guide the company's investments in technology as it competes with Apple, Google, and Amazon. ‘it always bothered me that we confused an enduring mission with a temporal goal,’ he said.  \\

\bottomrule
\end{tabular}
\vspace{-6pt}
\end{table*}

\begin{table*}[h]
    \centering
    \small
    \begin{tabular}{p{\columnwidth}@{}l@{}}
\toprule
\textbf{Document} 
 {\color{cyan} Officials have discovered that Islamic State jihadis have been using a chemistry laboratory at Mosul University to make bombs used by ISIS jihadists throughout the region.} “The University of Mosul is the best Daesh research center in the world,” Gen. Hatem Magsosi, Iraq’s main explosives officer, told The Wall Street Journal. “Trainees go to Raqqa, [Syria] then to Mosul university to use the existing facilities. ”  ISIS hijacked university chemistry lab in Mosul for making bombs: https:  .  by @MargaretWSJ @BKesling pic. twitter.   —   WSJ Think Tank (@WSJThinkTank) April 1, 2016,  { \color{magenta} They have found “  chemical bombs and suicide bomb vests like the ones used in the Brussels attacks and by at least some of the Paris attackers. ”} { \color{cyan} The lab also contained “  explosives and chemical weapons. ” However, officials told the outlet they do not know how much of the facility remains intact currently. The United   coalition bombed the university in March.} Alumni said the university boasted “a strong reputation around Iraq for its science departments. ” A year ago, the Islamic State established “a research hub in the chemistry lab. ” The terrorist group kept the staff at the university, many who “specialized in organic, industrial and analytical chemistry. ” A raid in Syria in March killed Islamic State’s    Abd   Mustafa   also known as Haji Imam. He taught physics in Iraq before he joined   in 2004. Officials put him in prison, but released him in 2012. Then he traveled to Syria, where he eventually joined the Islamic State. Gen. Magsosi said the group places Imam as “the top expert at the Mosul bomb lab. ” The sources told the Journal that the Islamic State used one part of the university for explosives and another for suicide bombs. The Wall Street Journal reports: During the same time frame, there has been a surge in Islamic State’s use of bombs that mix chemical precursors into an explosive powdery substance known as triacetone triperoxide, or TATP, both in Iraq and Europe. It isn’t clear how many of these weapons, if any, can be traced to research or training conducted in Mosul. Gen. Magsosi says that his   units called   explosives the “Satan Recipe” because they are very hard to detect and they are usually so lethal. The Islamic State captured Mosul, Iraq’s second largest city, in June 2014. Since then, they have destroyed libraries and buildings at the university. Kurdish outlet Rudaw reported last October that the group destroyed the university’s Faculty of Agriculture buildings. { \color{orange} In December 2014, ISIS raided the Central Library of Mosul to destroy all   books. ‘These books promote infidelity and call for disobeying Allah,” announced a militant to the residents. “So they will be burned. ” The library was “the biggest repository of learning the northern Iraqi town. ” The terrorists destroyed “Iraq newspapers dating to the early 20th century, maps and books from the Ottoman Empire, and book collections contributed by about 100 of Mosul’s establishment families. ” After that raid, the ISIS militants targeted the library at the University of Mosul. They burned science and culture textbooks in front of the students.} According to the Boston Globe: A University of Mosul history professor, who spoke on condition he not be named because of his fear of the Islamic State, said the extremists started wrecking the collections of other public libraries last month. { \color{orange} He reported particularly heavy damage to the archives of a Sunni Muslim library, the library of the    Latin Church and Monastery of the Dominican Fathers, and the Mosul Museum Library with works dating back to 5000 BC.} Citing reports by the locals who live near these libraries, the professor added that the militants used to come during the night and carry the materials in refrigerated trucks with   license plates. Militants also targeted the public library, which was home to more than 8, 000 rare books and manuscripts. Elderly residents begged the men not to burn the building.
 \\ \midrule
\textbf{Aspect:} { \color{cyan} chemistry} \\
\textbf{Summary:} Chemistry lab at Mosul University used to make bombs used by ISIS jihadists throughout the region, officials say. Officials say they do not know how much of the facility remains intact currently after coalition bombed the university in March. ISIS used one part of the university for explosives and another for suicide bombs, sources say.  \\ \midrule

\textbf{Aspect:} { \color{magenta} europe} \\
\textbf{Summary:} Officials have found chemical bombs and suicide bomb vests like the ones used in the Brussels attacks and by at least some of the Paris attackers. the lab also contained “  explosives and chemical weapons. It is not clear how many of these weapons, if any, can be traced to research or training conducted in the university. \\ \midrule

\textbf{Aspect:} { \color{orange} book} \\
\textbf{Summary:} Book collections destroyed in front of students at university library in Mosul, Iraq. ISIS has destroyed libraries and buildings at the university since it captured the city in June 2014, including one with works dating back to 5000 BC. Officials say they do not know how much of the facility remains intact currently. \\

\bottomrule
\end{tabular}
\vspace{-6pt}
\end{table*}

\end{document}